\documentclass[runningheads]{llncs}
\usepackage[T1]{fontenc}
\usepackage{graphicx}
\usepackage{float} 
\usepackage{amsfonts,amssymb}
\usepackage{amsmath}
\usepackage{array}  
\usepackage{tabularx} 
\usepackage{hyperref}
\begin{document}
\title{DuoFormer: Leveraging Hierarchical Visual Representations by Local and Global Attention}
\titlerunning{DuoFormer: Leveraging Hierarchical Visual Representations}

\author{Xiaoya Tang\inst{1}\orcidID{0009-0002-5638-320X} \and
Bodong Zhang\inst{1,2}\orcidID{0000-0001-9815-0303} \and
Beatrice S. Knudsen\inst{3}\orcidID{0000-0002-7589-7591} \and\\
Tolga Tasdizen\inst{1,2}\orcidID{0000-0001-6574-0366}}

\authorrunning{X. Tang et al.}

\institute{Scientific Computing and Imaging Institute, University of Utah, SLC, UT, USA \\
\email{xiaoya@sci.utah.edu}
\and
Electrical and Computer Engineering, University of Utah, SLC, UT, USA
\email{bodong.zhang@utah.edu}, \email{tolga@sci.utah.edu}\and Department of Pathology, University of Utah, Salt Lake City, UT, USA
\email{beatrice.knudsen@path.utah.edu}}
%
\maketitle              
\begin{abstract}
We here propose a novel hierarchical transformer model that adeptly integrates the feature extraction capabilities of Convolutional Neural Networks (CNNs) with the advanced representational potential of Vision Transformers (ViTs). Addressing the lack of inductive biases and dependence on extensive training datasets in ViTs, our model employs a CNN backbone to generate hierarchical visual representations. These representations are then adapted for transformer input through an innovative patch tokenization. We also introduce a 'scale attention' mechanism that captures cross-scale dependencies, complementing patch attention to enhance spatial understanding and preserve global perception. Our approach significantly outperforms baseline models on small and medium-sized medical datasets, demonstrating its efficiency and generalizability. The components are designed as plug-and-play for different CNN architectures and can be adapted for multiple applications. The code is available at https://github.com/xiaoyatang/DuoFormer.git.
\keywords{Vision Transformer  \and Inductive Bias \and Multi-scale features.}
\end{abstract}
\section{Introduction}
The Vision Transformer (ViT)~\cite{dosovitskiy2020image} has significantly advanced the adaptation of transformers from language to vision, demonstrating superior performance over CNNs when pre-trained on large datasets. ViT employs a patch tokenization process that converts images into a sequence of uniform token embeddings. These tokens undergo Multi-Head Self-Attention (MSA), transforming them into queries, keys, and values that capture extensive non-local relationships. Despite their potential, ViTs can underperform similarly-sized ResNets~\cite{he2016deep} when inadequately trained due to their lack of inductive biases such as translation equivariance and locality\cite{raghu2021vision,lee2021vision}, which are naturally encoded by CNNs. Recent efforts have focused on mitigating ViTs' limitations by integrating convolutions or adding self-supervised tasks. Prevalent approaches combine CNN feature extractors with transformer encoders \cite{araujo2019computing,dosovitskiy2020image,wu2021cvt,liu2021swin,yuan2021incorporating,fan2021multiscale,li2021localvit,d2021convit}, such as the 'hybrid' ViT~\cite{dosovitskiy2020image}. Other methods like knowledge distillation~\cite{touvron2021training} transfer biases from CNNs to ViT. Nonetheless, ViTs' smaller receptive fields compared to CNNs limit their ability to capture detailed spatial relationships~\cite{araujo2019computing}, which can be partially alleviated by techniques like enriched spatial shifting patches~\cite{lee2021vision}.

Histopathology image analysis, critical in medical diagnostics, involves examining whole slide images (WSIs) to detect and interpret complex tissue structures and cellular details. This analysis faces challenges due to the varied scales of visual entities within WSIs,  such as the differing sizes of cell nuclei and vascular structures, both of which can contribute to a model's task of distinguishing low- and high-risk kidney cancers. Moreover, vital global features of cancer and its microenvironment, observable only at lower scales, are crucial for various downstream tasks. The neglect of these multiple scales can significantly impair the performance of deep learning models in medical image recognition tasks. CNNs tackle this issue by utilizing a hierarchical structure created by lower and higher stages, which allows them to detect visual patterns from simple low-level edges to complex semantic features. Conversely, ViTs employ fixed-scale patches, thereby overlooking crucial multi-scale information within the same image~\cite{raghu2021vision}, which can hinder their performance across diverse tasks. By harnessing a hierarchical structure similar to that of CNNs, ViTs can be prevented from overlooking the critical multi-scale features, while also imparting necessary inductive biases. Existing works on directly integrating multi-scale information into ViTs vary primarily in the placement of convolutional operations: during patch tokenization\cite{yuan2021incorporating,xu2021co,guo2022cmt}, within\cite{hou2024conv2former,guo2022cmt,lin2023scale} or between self-attention layers, including query/key/value projections\cite{wu2021cvt,yuan2021incorporating}, forward layers~\cite{li2021localvit}, or positional encoding~\cite{xu2021co}, etc. Despite the benefits of hierarchical configurations~\cite{heo2021rethinking}, a definitive model for visual tasks has yet to emerge. The challenge remains in effectively producing and utilizing features across various scales. In response, we propose a novel hierarchical Vision Transformer model, outlined as follows:
\begin{itemize}
    \item[1.]Our proposed multi-scale tokenization involves a single-layer projection, patch indexing, and concatenation, assembling features from different stages of the CNN into multi-scale tokens.
    
    \item[2.]We introduce a novel MSA mechanism called scale attention, combined with patch attention. This approach enables the model to recognize connections across scales, expanding ViT's receptive field and bridging the gap between CNN and Transformer architectures.

    \item[3.]Our proposed scale token, part of the scale attention, is initialized with a fused embedding derived from hierarchical representations. It enriches the transformer's multi-granularity representation and aggregates scale information, serving as the input for the global patch attention.
\end{itemize}
\section{Related Work}
Various approaches have explored integrating the hierarchical architecture of CNNs into Vision Transformers (ViTs) across different visual tasks, including video recognition\cite{fan2021multiscale}, image classification \cite{chen2021crossvit,dong2022cswin,d2021convit,fan2024rmt,heo2021rethinking,hou2024conv2former,li2021localvit,lin2023scale,liu2021swin,touvron2021training,wang2021pyramid,wang2023crossformer++,wu2021cvt,xu2021co,yu2022metaformer}\\
\cite{yuan2021incorporating,yuan2021tokens,zhang2024mg}, object detection\cite{dong2022cswin,fan2024rmt,heo2021rethinking,hou2024conv2former,lin2023scale,wang2021pyramid,xia2024vit,xu2021co,yu2022metaformer,yuan2021incorporating,zhang2024mg}, and segmentation\cite{dong2022cswin,fan2024rmt}\\
\cite{hou2024conv2former,lin2023scale,wang2022uctransnet,wang2021pyramid,xu2021co,yu2022metaformer,zhang2024mg}. Notable methods emulate the pyramid structure of CNN with stage-wise pooling and convolutional embeddings~\cite{heo2021rethinking} or integrate pooling within the attention mechanism\cite{fan2021multiscale}.

Multiple scales have been exploited beyond mere convolution integration. The Swin Transformer~\cite{liu2021swin} utilizes a shifting window strategy, while Dong et al.~\cite{dong2022cswin} split multi-heads to perform self-attention in horizontal and vertical stripes. Chen et al.~\cite{chen2021crossvit} developed a dual-branch architecture that processes varying patch sizes, and Zhang et al.~\cite{zhang2024mg} implemented a multi-granularity strategy. PVT~\cite{wang2021pyramid} reduces feature size progressively using spatial-reduction attention. A recent study~\cite{fan2024rmt} employed a spatial decay matrix to enhance self-attention with spatial priors. UNETR~\cite{hatamizadeh2022unetr} constructs a U-shape transformer encoder and decoder for 3D segmentation, while people also replaced UNet skip connections with attention mechanisms\cite{wang2022uctransnet} for 2D segmentation. Besides, inductive biases can also be integrated through auxiliary tasks such as unsupervised localization to enhance local processing capabilities~\cite{liu2021efficient}.

\section{Methodology}
\begin{figure}[h]
\centering
\includegraphics[width=0.75\linewidth]{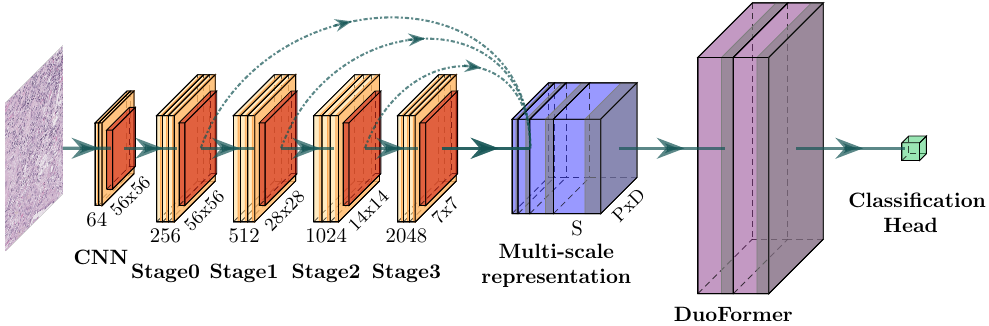} 
\caption{The pipeline of the proposed DuoFormer. Dimensionalities of the multi-scale representation: S: scale dimension; P: number of patches; D: embedding dimension. } \label{fig:1}
\end{figure}

Our model utilizes a CNN as the embedding layer, depicted in Figure \ref{fig:1}. Patch tokenization contains two steps represented by the dashed lines in Figure \ref{fig:1}: First is extracting hierarchical representations from different stages of the CNN backbone. Second is the projection and patch indexing.  After acquiring the multi-scale features, we use our DuoFormer to learn the local dependencies across scales and global dependencies across patches, which are needed for downstream tasks. 
\begin{figure}[h]
\centering
\includegraphics[width=0.8\linewidth]{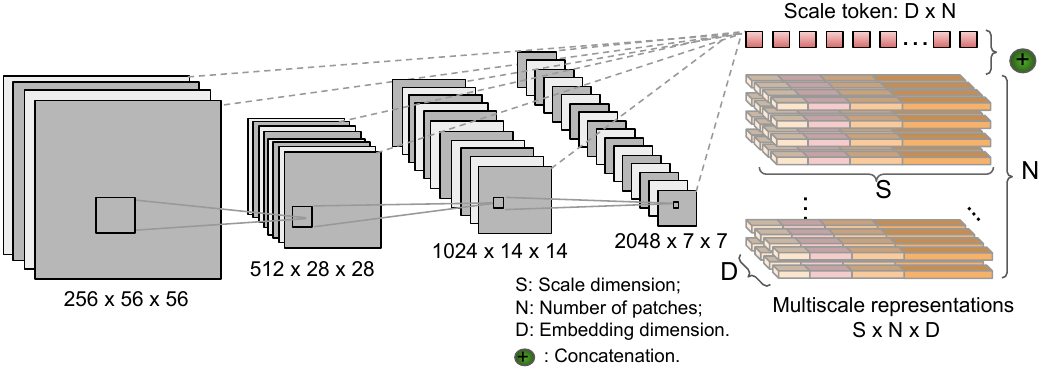} 
\caption{Visualization of Multiscale Patch Tokenization: This figure depicts the process of converting an image into a sequence of multi-scale patch embeddings, with each color representing a different scale to illustrate the varied dimensions of the patches.} 
\label{fig:2}
\end{figure}
\subsection{Multi-scale Patch Tokenization}
Given the input to the backbone, $\mathbf{x} \in \mathbb{R}^{H \times W \times 3}$ with $H=W$, we derive hierarchical outputs from four stages, denoted as $\textbf{x}_i \in \mathbb{R}^{P_i \times P_i \times C_i}$ for $i \in {0,1,2,3}$. Here, $P_i = \frac{H}{4 \cdot 2^i}$ specifies the spatial resolution, and $C_i$ represents the channel dimension. We apply a linear projection to transform these features into embeddings of dimension $D$. Next, we split the embeddings $\mathbf{x}'$ from each stage into $N$ non-overlapping patches, set as $N=49$. Each scale yields a sequence of flattened tokens with spatial size ${P'_i}^2$, where ${P'_i} = \frac{H}{4 \cdot 2^i \cdot \sqrt{N}}$. We index and concatenate multi-scale embeddings from each patch across all scales to form the multi-scale tokens $\textbf{X}^t_{\sum}$, illustrated in Figure \ref{fig:2}. The equation for this process is:
\begin{equation}
    \begin{aligned}
    \mathbf{x}'_i &= \text{Projection}(\textbf{x}_i), \mathbf{x}'_i \in \mathbb{R}^{P_i \times P_i \times D}\\
    \mathbf{x}''_i &\in \mathbb{R}^{{P'_i}^2 \times N \times D}, {P'_i}^2 = \frac{HW}{16 \cdot 4^i \cdot N}, i \in {0, 1, 2, 3}, \\
    \mathbf{X}^t_{\sum} &= \textbf{concat}(\mathbf{x}''_i) \in \mathbb{R}^{S \times N \times D}, S = \sum {P'_i}^2.
    \end{aligned}
\end{equation}

\begin{figure}[h]
\centering
\includegraphics[width=0.75\linewidth]{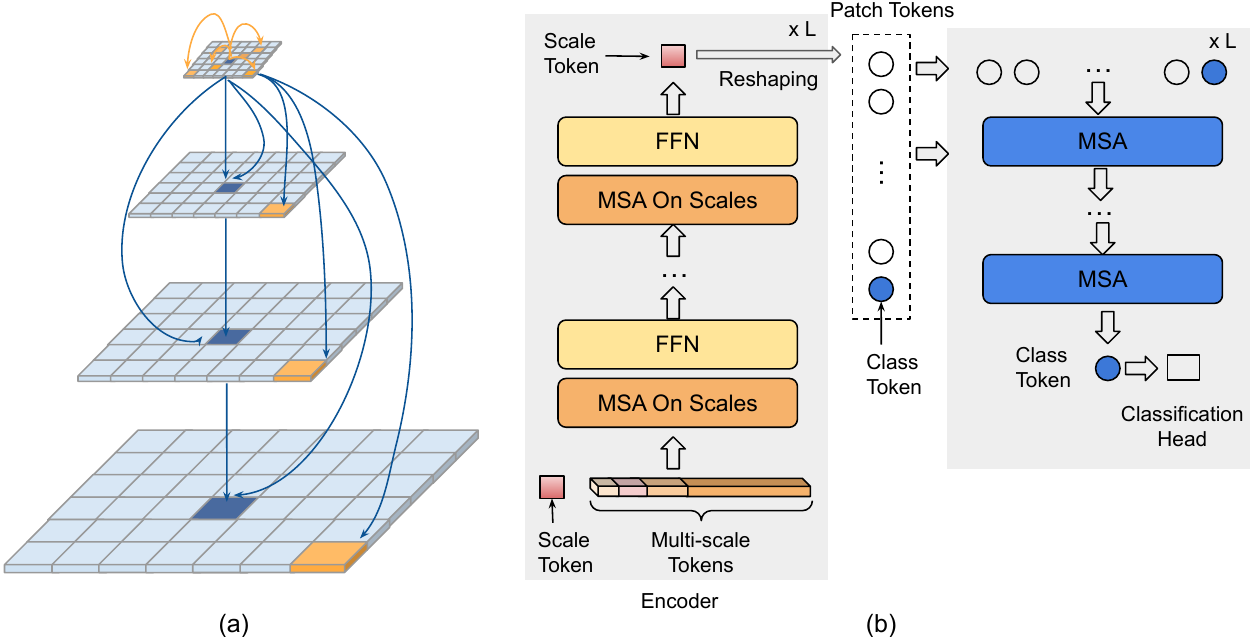} 
\caption{Illustration of the Duo attentions. Panel (a) shows the local (yellow arrows) and global (blue arrows) dependencies among multi-scale patches, maintaining a consistent grid size of 49; larger patches indicate greater embedding lengths. Panel (b) details the model architecture, including L layers of scale and patch attention blocks in the encoder. }
\label{fig:3}
\end{figure}
\subsection{Duo Attention Module}
Our tokenization directly embeds multiscale spatial information into the scale dimension, inherently enriching the model's inductive biases. Subsequently, our encoder employs scale and patch attentions to respectively focus on detailed image features and broader contexts, as illustrated in Figure \ref{fig:3}(a). Our scale attention adapts the Multi-Head Self-Attention (MSA) framework by incorporating an additional scale dimension. This adaption integrates multi-scale analysis directly into the attention mechanism and alters tensor operations to accommodate multi-dimensional tokens. Details are explained in the equation below and depicted in Figure \ref{fig:3}(b). 
\begin{equation}
\mathbf{X}_{\sum}^\prime = \mathbf{X}_{\sum} + \text{MSA}(\text{LN}(\mathbf{X}_{\sum})), \quad \mathbf{Y} = \mathbf{X}_{\sum}^\prime + \text{FFN}(\text{LN}(\mathbf{X}_{\sum}^\prime))
\end{equation}
After scale attention, the scale token aggregates key details from all scales for input to patch attention. Patch attention, mirroring standard MSA, omits layer normalization(LN), feed-forward networks (FFN), and residual connections, as shown in Figure \ref{fig:3}(b).
\subsection{Scale Token}
To enhance the hierarchical representations, we use a downsampling strategy involving simple convolutional layers followed by max pooling. This process normalizes the spatial dimensions of all embeddings to $N$, maintaining consistent channel dimensions. $N$ denotes number of patches, set as 49 in our experiments. These embeddings are then concatenated along the channel dimension and projected into the embedding dimension $D$ using lightweight convolutions. The resulting scale token, concatenated with multi-scale tokens, serves as the input for the scale attention, guiding it effectively, as detailed below.
\begin{equation}
\begin{aligned}
\Tilde{\textbf{x}}_0 &= \text{MaxPool(Conv(}\textbf{x}_0\text{))}, \quad \Tilde{\textbf{x}}_1 = \text{MaxPool(Conv(}\textbf{x}_1\text{))} \\
\Tilde{\textbf{x}}_2 &= \text{MaxPool(}\textbf{x}_2\text{)}, \quad \Tilde{\textbf{x}}_3 = \textbf{x}_3 ,\quad\text{where } \textbf{x}_i \in \mathbb{R}^{N \times C_i},\\
\Tilde{\mathbf{X}}_{\sum} &= \textbf{concat}(\Tilde{\textbf{x}_0}, \Tilde{\textbf{x}_1}, \Tilde{\textbf{x}_2}, \Tilde{\textbf{x}_3}) \in \mathbb{R}^{N \times C}, C = \sum C_i, \\
\text{Scale Token} &= \text{ReLU(BN(Conv(}\Tilde{\mathbf{X}}_{\sum}\text{)))} \in \mathbb{R}^{N \times D}
\end{aligned}
\end{equation}

\section{Experiments}
\subsection{Experimental Setup}
Our evaluation utilized two datasets, Utah ccRCC and TCGA ccRCC ~\cite{zhang2023class}, with varying ResNet backbones for a thorough analysis. The Utah ccRCC dataset comprises 49 WSIs from 49 patients, split into training (32 WSIs), validation (10 WSIs), and testing (7 WSIs). Tiles were extracted from marked polygons at 400x400 pixel resolution at 10X magnification with a 200-pixel stride and center-cropped to 224x224 pixels for model compatibility. The training set included 28,497 Normal/Benign, 2,044 Low Risk, 2,522 High Risk, and 4,115 Necrosis tiles, with validation and test sets proportionately distributed. The TCGA ccRCC dataset features 150 labeled WSIs divided into 30 for training, 60 for validation, and 60 for testing, using similar cropping methods but adjusted strides to gather more training patches. It contains 84,578 Normal/Benign, 180,471 Cancer, and 7,932 Necrosis tiles in the training set, with similar distributions in validation and test sets. For model details, please refer to Appendix.

\subsection{Result Analysis}
In this study, we utilized ResNet18 and ResNet50 backbones~\cite{he2016deep} pre-trained on extensive datasets, assessing our model under two paradigms: with ImageNet supervised pre-trained and pathology(TCGA and TULIP) self-supervised pre- trained~\cite{kang2023benchmarking} backbones. Results, shown in Table \ref{tab1}, indicate our model surpasses ResNet baselines by over 2\% across all settings and outperforms various Hybrid-ViTs in all scenarios. The results underscore our model's capacity to harness multi-scale features and integrate crucial inductive biases without necessitating additional tasks or additional pre-training of the transformer encoder.

In the supervised pre-training scenario, particularly with TCGA using a ResNet 50 backbone, deeper encoders sometimes hindered performance, highlighting the need for careful design when integrating CNN architectures, especially considering domain shifts. Our DuoFormer improved performance by 3.83\%, demonstrating its effectiveness in leveraging multi-scale representations and likely guiding the feature extractor to adapt better to domain shifts when trained together. In the self-supervised pre-trained experiments, our model significantly outperformed the baseline by 9.88\% and clearly surpassed the Hybrid-ViTs, showing the superiority of our model in leveraging multi-scale features. These findings suggest that with the proposed designs, the model can effectively capture essential local features while preserving global attention capabilities, thereby addressing the typical inductive bias limitations found in transformers.

\begin{table}[t]
\centering
\caption{
Feature extractors were unfrozen unless specified otherwise. Results here are reported as mean values from several independent experiments.}
\label{tab1}
\begin{tabularx}{\textwidth}{|>{\hsize=1.8\hsize}X|>{\centering\arraybackslash\hsize=0.5\hsize}X|>{\centering\arraybackslash\hsize=0.7\hsize}X|} 
\hline
\multicolumn{3}{|c|}{\textbf{ImageNet Supervised Pretrained}} \\ \hline
\textbf{Model} & \textbf{Params.} & \textbf{Accuracy (\%)} \\
\hline
\multicolumn{3}{|c|}{\textbf{TCGA}} \\
\hline
ResNet50 & 23.50M & $72.74$ \\
ResNet50-ViT Base & 112.5M & $75.89$ \\
ResNet50-ViT Large & 197.6M & $73.34$ \\
ResNet50-DuoFormer (Ours) & 186.0M & \textbf{76.57} \\ \hline
\multicolumn{3}{|c|}{\textbf{UTAH}} \\
\hline
ResNet18 & 11.20M & $88.87$ \\
ResNet18-ViT Base & 99.03M & $82.35$ \\
ResNet18-ViT Large & 184.1M & $86.39$ \\
ResNet18-DuoFormer (Ours) & 89M & \textbf{91.22} \\
\hline
\multicolumn{3}{|c|}{\textbf{Pathology Self Supervised Pretrained  }} \\ \hline
\multicolumn{3}{|c|}{\textbf{TCGA}} \\ \hline
\textbf{Model} & \textbf{Params.} & \textbf{Accuracy (\%)} \\
\hline
ResNet50-SwaV (Freeze) & 0.008M & $77.98$ \\
ResNet50-SwaV (Freeze)-ViT Base & 89.03M & $74.00$ \\
ResNet50-SwaV (Freeze)-ViT Large & 174.1M & $85.81$ \\
ResNet50-SwaV (Freeze)-DuoFormer (Ours) & 124.7M & \textbf{86.45} \\
\hline
\end{tabularx}
\end{table}

\subsection{Ablation Studies}
\subsubsection{Ablation on Scale Attention}
For ablation studies, we utilized our best models in both settings, employing ResNet18 pretrained on ImageNet for UTAH and ResNet50 pretrained on histopathology images for TCGA. We evaluated the individual contributions of scale and patch attention mechanisms using configurations of 6 layers for UTAH and 8 layers for TCGA. The different depths were chosen to adapt to the larger size of the TCGA dataset and the smaller size of the UTAH dataset. Results in Table \ref{tab2} indicate that scale attention alone outperforms setups using only patch attention, suggesting the robustness of our scale token and attention mechanism in harnessing multi-scale features. The results also confirm that the best performance on both datasets is achieved when both attention modules are employed together, emphasizing the necessity of integrating both local and global information for effective visual processing. 
\begin{table}[t]
    \centering
    \caption{Ablation study comparing scale and patch attention individually and in combination. Configurations with only scale attention use a single fully-connected (FC) layer to adapt the scale token for the classification head, trained alongside the entire network.}
    \label{tab2}
    \begin{tabular}{|c|c|c|c|}
        \hline
        \textbf{Dataset} & Scale Attn & Patch Attn & Scale Attn \& Patch Attn ({Ours}) \\
        \hline
        \textbf{UTAH} & \multicolumn{1}{c|}{$90.31$} & \multicolumn{1}{c|}{$82.35$} & \multicolumn{1}{c|}{$\mathbf{91.22}$} \\
        \hline
        \textbf{TCGA} & \multicolumn{1}{c|}{$79.90$} & \multicolumn{1}{c|}{$74.00$} & \multicolumn{1}{c|}{$\mathbf{86.45}$} \\
        \hline
    \end{tabular}
\end{table}

\subsubsection{Ablation on Scale Token}
To evaluate the role of the scale token, we conducted experiments comparing configurations with and without a scale token, and against a learnable scale token, as shown in Table \ref{tab3}. Our scale token effectively enhanced local information capture, outperforming the learnable version. Without a scale token, using the first token from scale attention yielded better results than averaging all tokens, likely due to the first token's representation of the final CNN stage's output, which provides crucial, concise information. This suggests that averaging introduces noise. 
\begin{table}[t]
    \centering
    \caption{Ablation study on the impact of different scale token configurations, including a learnable scale token implemented as nn.Parameters(). }\label{tab3}
    \begin{tabular}{|c|c|c|c|c|}
        \hline 
        \textbf{Dataset} & The first token & Avg. of tokens & Learnable & Ours \\
        \hline
        \textbf{} & w/o Scale-Token & w/o Scale-Token & w/i Scale-Token & w/i Scale-Token \\
        \hline
        \textbf{UTAH} & $90.61$ & $89.62$ & $88.80$ & $\mathbf{91.22}$ \\
        \hline
        \textbf{TCGA} & $83.22$ & $82.62$ & $83.13$ & $\mathbf{86.45}$\\
        \hline
    \end{tabular}
\end{table}
\subsubsection{Ablation on Multi-Scale Representations}
We explored the impact of different combinations of stages on the UTAH dataset. $S_0$ represents the shallowest stage ($56\times 56$), and $S_3$ is the deepest ($7\times 7$). According to the results, incorporating all stages slightly harmed performance, likely due to overfitting given the small UTAH dataset. Including $S_3$ generally improved performance, highlighting the final stage's importance for classification accuracy. Including $S_0$ often decreased performance, possibly due to its larger spatial embeddings and higher overfitting risk. Conversely, on larger datasets, as shown in Table \ref{tab1}, including all stages proved beneficial. The highest three configurations here used $S_1$, $S_2$, and $S_3$, demonstrating the benefits of multi-scale integration while managing computational complexity. 
\begin{figure}[H]
\centering
\includegraphics[width=0.7\linewidth,height=0.23\textheight]{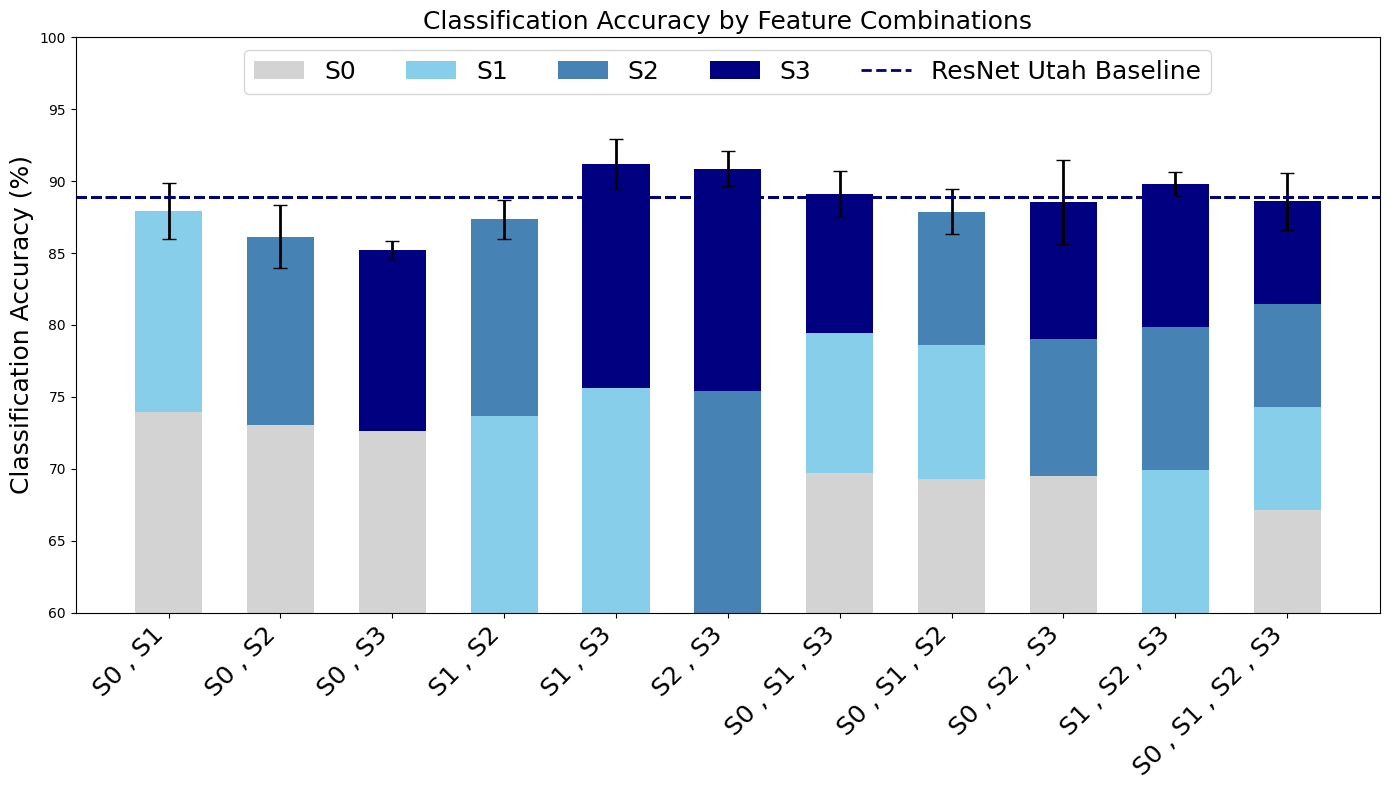} 
\caption{Ablation study on combinations of hierarchical stages. Stages are represented by colors from light to dark. Bar heights and black error bars show mean accuracies and standard deviations, and the blue dashed line marks the ResNet baseline.} \label{fig:4}
\end{figure}


\section{Conclusion}
In this study, we introduced a novel hierarchical transformer with dual attention mechanisms that enhance visual data interpretation across scales, improving medical image classification. Ablation studies confirm performance optimization, demonstrating the model's robustness and adaptability across various CNN backbones and tasks, paving the way for broader applications in medical imaging and vision-related fields.




\bibliographystyle{splncs04}
\bibliography{DuoFormer}

\newpage
\section{Appendix}
\subsubsection{Model Training Details}
All models, including the ResNet baselines, were trained using the Adam optimizer with \(\beta_1 = 0.9\) and \(\beta_2 = 0.999\), without applying weight decay. For the DuoFormer model, batch sizes were set to 32 for the Utah dataset and 6 for the TCGA dataset. We employed a OneCycle learning rate scheduler that starts from a minimal learning rate, progressively increasing to a set rate of \(1 \times 10^{-4}\). Each model underwent training for 50 epochs on Utah and 100 epochs on TCGA, utilizing early stopping with patience of 20 and 50 epochs, respectively. We saved the best-performing model from the validation data for inference. Model performance was evaluated using balanced accuracy for both datasets. All computations were performed on an NVIDIA RTX A6000 with 48 GB of memory.


\subsubsection{Ablation on Numbers of Heads and Layers}
We assessed our model's sensitivity to two hyperparameters: the number of heads and the number of layers in two attention modules. Initially, we fixed the number of heads at 12 and varied the number of layers from 4 to 12 to identify optimal configurations for each dataset. Subsequently, we tested heads from 4 to 12, excluding 10 due to incompatibility with the feature dimension $D=768$, using the optimal number of layers. We observed that performance generally increases and then decreases with attention depth. Specifically, performance peaks at 6 layers for the Utah dataset and at 8 layers for the TCGA dataset, likely due to the varying sizes of the datasets. Additionally, we noted a similar pattern of initial increase followed by a decrease in performance for the number of heads across both datasets, peaking at 8 heads. Notably, our models with all tested numbers of heads and layers performed better than baseline ResNets, except in one case where performance was slightly worse, demonstrating the effectiveness of our proposed model.

\begin{figure}[H]
\centering
\includegraphics[width=0.78\linewidth,height=0.18\textheight]{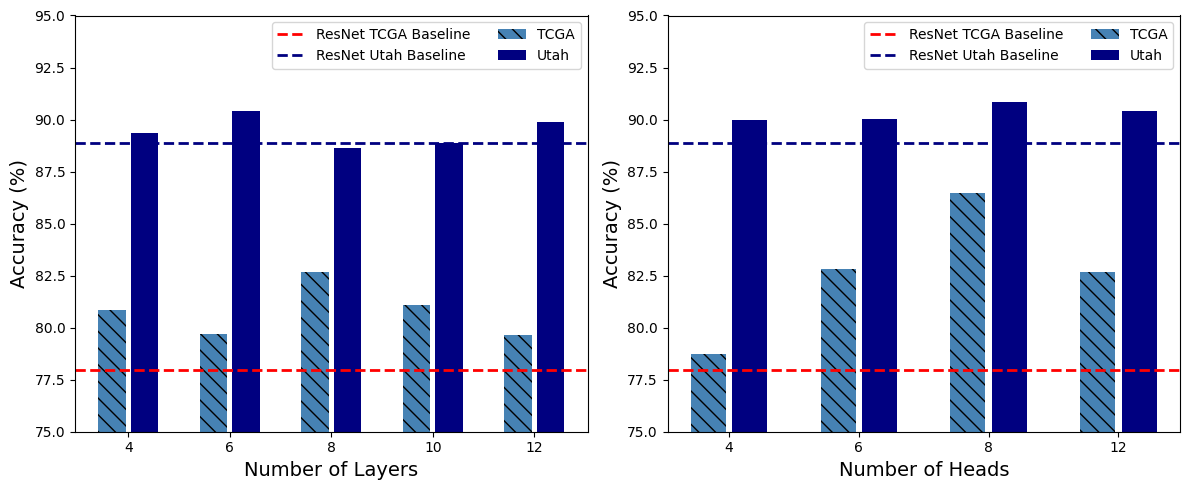} 
\caption{Ablation studies comparing the number of layers and heads in the dual attention modules for both the TCGA (solid bars) and Utah (striped bars) datasets. The dashed lines represent ResNet baselines for each dataset. Each configuration synchronizes the layers between scale and patch attention.}
\label{fig:5}
\end{figure}
\end{document}